\pdfoutput=1
\documentclass[11pt]{article}
\usepackage{ACL2024}
\usepackage{times}
\usepackage{latexsym}
\usepackage[T1]{fontenc}
\usepackage[utf8]{inputenc}
\usepackage{microtype}
\usepackage{inconsolata}
\usepackage{booktabs}
\usepackage[most]{tcolorbox}
\usepackage{listings}
\usepackage{color}
\usepackage[inline]{enumitem}
\usepackage{twemojis}

\hypersetup{colorlinks, citecolor=teal, linkcolor=red, urlcolor=blue}

\definecolor{dkgreen}{rgb}{0,0.6,0}
\definecolor{gray}{rgb}{0.5,0.5,0.5}
\definecolor{mauve}{rgb}{0.58,0,0.82}

\usepackage{amsfonts}
\usepackage{graphicx} 
\usepackage{multirow}
\usepackage{caption}
\usepackage{subcaption}

\usepackage{wrapfig, lipsum, booktabs}
\usepackage{amsmath}
\newcommand{\context}{\text{\small\textsf{context}}}
\newcommand{\scriptcontext}{\text{\scriptsize\textsf{context}}}
\newcommand{\query}{\text{\small\textsf{query}}}
\newcommand{\answer}{\text{\small\textsf{answer}}}
\newcommand{\cache}{\text{\small\textsf{cache}}}

\newcommand{\kvcache}{KV}
\newcommand{\jitkvcache}{JIT-KV}
\newcommand{\QA}{QA}

\newcommand{\Oh}[1]{\mathcal{O}\bigl({#1}\bigr)}
\newcommand{\Ohlinear}{\Oh{\lvert\context\rvert}}
\newcommand{\Ohquad}{\Oh{\lvert\context\rvert^2}}  
\newcommand{\Ohcacheinfer}{\Oh{\lvert\context\rvert (\lvert\query\rvert + \lvert\answer\rvert)}}  
\newcommand{\Ohfull}{\Oh{(\lvert\context\rvert + \lvert\query\rvert + \lvert\answer\rvert)^2}}

\newcommand{\ie}{\textit{i.e.}}
\newcommand{\eg}{\textit{e.g.}}

\newcommand{\xccache}{{\textsc{XC-Cache}}}
\newcommand{\xc}{{\textsc{XC}}}
\newcommand{\ourmodel}{{\textsc{XC-Llama}}}
\newcommand{\ourmodelencoder}{{\textsc{XC-LlamaEnc}}}
\newcommand{\llama}{{\textsc{Llama~2}}}
\newcommand{\llamachat}{{\textsc{Llama~2-Chat}}}

\newcommand{\hotpotqa}{\textsc{HotpotQA}}

\newcommand{\nq}{\textsc{NQ}}

\usepackage{pifont}
\def\cmark{{\ding{51}}}%
\def\xmark{{\ding{55}}}%

\title{\xccache{}: Cross-Attending to Cached Context for Efficient LLM Inference}

\author{João Monteiro\textsuperscript{\textdagger,3} Étienne Marcotte\textsuperscript{\textdagger,1}, Pierre-André Noël\textsuperscript{\textdagger,1}, Valentina Zantedeschi\textsuperscript{\textdagger,1},\\ \textbf{David Vázquez\textsuperscript{1}, Nicolas Chapados\textsuperscript{1}, Christopher Pal\textsuperscript{1,2}, Perouz Taslakian\textsuperscript{\textdagger,1}} \\
  1-ServiceNow Research\\
  2-Québec Artificial Intelligence Institute\\
  3-Autodesk -- Work done while at ServiceNow\\
  \textsuperscript{\textdagger}Core contributors}

\begin{document}
\maketitle

\begin{abstract}
Prompts are often employed to condition decoder-only language model generation on reference information. Just-in-time processing of a context is inefficient due to the quadratic cost of self-attention operations, and caching is desirable.
However, caching transformer states can easily require almost as much space as the model parameters. When the right context is not known in advance, caching the prompt can be challenging. This work addresses these limitations by introducing models that, inspired by the encoder-decoder architecture, use cross-attention to condition generation on reference text without the prompt. 
More precisely, we leverage pre-trained decoder-only models and only train a small number of added layers. We use Question-Answering (QA) as a testbed to evaluate the ability of our models to perform conditional generation and observe that they outperform prompt-based inference methods, are comparable to fine-tuned prompted LLMs, and drastically reduce the space footprint relative to standard \kvcache{} caching by two orders of magnitude. Specifically, we introduced \ourmodel{} which converts a pre-trained \llama{} into an encoder-decoder architecture by integrating cross-attention layers interleaved in between existing self-attention layers.
\end{abstract}

\section{Introduction}

Large Language Models (LLMs) have propelled advances in language modeling and enabled automatic production of almost human-like text. 
Despite impressive progress, challenges persist in applying LLMs in practical settings such as the risk of \emph{hallucinations} (or rather \emph{confabulatations}~\citep{bottou2023borges, berenblog}) and of non-factual~\citep{li2023halueval, xu2024hallucination} or toxic content~\citep{zou2023universal, xhonneux2024context} in their generated text.
Moreover, without fine-tuning, it is surprisingly difficult to adapt these models to incorporate new information not included in their training data~\citep{luo2023empirical, kalajdzievski2024scaling}. 
Indeed, while LLMs can answer queries about their training data somewhat accurately in a zero-shot fashion~\citep{petroni2019language}, queries regarding information not available at training time often lead to inaccuracies~\citep{maynez2020faithfulness, ji2023survey} as one would expect. 

\begin{figure}[t]
    \centering
    \includegraphics[width=0.9\linewidth]{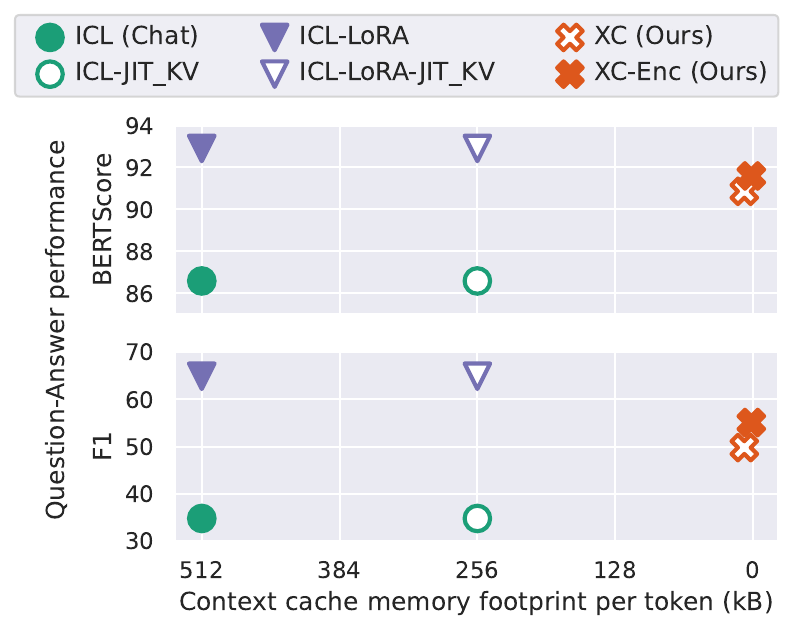}
    \caption{Average \QA{} performance vs. caching memory footprint per context token. \textbf{The closer to the top-right corner, the better}. \ourmodel{} variants drastically reduce cache size at a small cost in accuracy.}
    \label{fig:mem_bertscore}
\end{figure}

\begin{figure}[t]
\centering
\includegraphics[width=\linewidth]{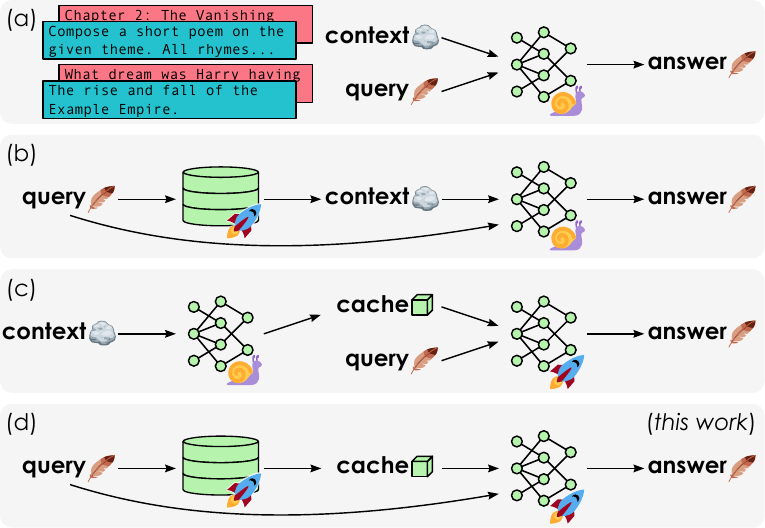}
\caption{%
\textbf{Faster inference in context-conditional language modeling.}
(a) A use case where the user's $\query$ must be interpreted within some $\context$ to generate an $\answer$.
This work focuses on cases where the $\query$ and the $\answer$ are both small (light \twemoji{feather}), but the $\context$ is large (heavy \twemoji{rock}). The resulting LLM time complexity is thus $\Ohquad$ (slow \reflectbox{\twemoji{snail}}).
(b) In-context learning (ICL) and retrieval augmented generation (RAG) are two examples where the $\query$ is used to look up the $\context$ \emph{from a finite corpus}.
(c) In many cases, the $\context$ can be processed in advance to a $\cache$ enabling $\Ohcacheinfer$ (fast \twemoji{rocket}) inference on a given $\query$.
(d) Finite $\context$ corpus may be processed offline, enabling fast execution at inference. Since $\cache$ size affects storage and communication costs, we search for models requiring smaller $\cache$.
}%
\label{fig:cond_gen_and_caching}
\end{figure}

This work focuses on grounding LLM generation on contextual information provided at inference time
(Figure~\ref{fig:cond_gen_and_caching}(a)).
The most common approach for conditioning model generation is arguably \emph{In-Context Learning}\footnote{Throughout this paper, we use \emph{in-context learning} to refer to a 0-shot prompt-based inference setting where the information required to solve the task is included in the prompt.} (ICL)~\citep{radford2019language, brown2020language}: one prepends the relevant context to the prompt to generate an answer conditioned on the combined query and context.
This technique is a core component of popular frameworks, such as Retrieval-Augmented Generation (RAG)~\citep{lewis2020retrieval}, with the specificity that the relevant context is not known \textit{a priori}, but has to be retrieved from a corpus~\citep{ram2023context}. 

While somewhat effective and straightforward, ICL, as typically performed with decoder-only architectures, has flaws. 
On the one hand, ICL-based generation is known to present high variance with respect to the prompt template so that equivalent valid-looking prompt formats produce drastically different results~\citep{chen2023relation}.
On the other hand, ICL is costly in terms of time and space. 
Just-in-time processing of the context suffers from quadratic complexity on the length due to self-attention operations~\cite{vaswani2017attention}.
The alternative is pre-processing and caching the context internal states (the so-called key-value or \kvcache{} states) to speed up inference (Figure~\ref{fig:cond_gen_and_caching}(c)). 
However, this \emph{can require the same order of space as the model parameters themselves} (we give details in Section~\ref{sec:caching}). 
Recent work has reduced the space requirements of \kvcache{} caching by sub-sampling states~\citep{xiao2023efficient, adnan2024keyformer}, although at the cost of ignoring relevant content. 

To overcome these limitations, we propose alternatives to ICL that perform conditional generation without injecting the relevant information in the prompt (Figure~\ref{fig:cond_gen_and_caching}(a)), and seek to implement lightweight cache methods as illustrated in Figure~\ref{fig:cond_gen_and_caching}(d).
Our approach is reminiscent of the, now arguably legacy, encoder-decoder architectures, as it relies on cross-attention layers to condition generation on pre-computed context encodings. More precisely, we propose cross-context-cache (\xccache{}), which stores only the outputs of the encoder hidden states and relies on cross-attention to ingest the cache at inference time.
We instantiate \xccache{} via two parameter-efficient approaches that leverage pre-trained decoder-only models and extend them with a separate encoder to process the context: one approach uses the frozen decoder as an encoder (called \ourmodel{}), and the other uses a small bi-directional encoder (called \ourmodelencoder{}). Crucially, our encoder-decoder architectures are more amenable to caching the context states, requiring orders of magnitude less space than their ICL counterparts. 
When context caching is enabled, fine-tuned models result in higher accuracy performance, but demand a large memory footprint (and consequently, higher latency and cost). In contrast, our \xccache{} approach substantially reduces cache memory requirements by nearly 98\%; and as other efficiency-improving techniques like quantization~\citep{frantar2023gptq}, this reduction comes at a minor cost in accuracy, as illustrated in Figure~\ref{fig:mem_bertscore}.
Nevertheless, our method consistently outperforms ICL alternatives based on \llama{} or \textsc{GPT-3.5}, as detailed in Section~\ref{sec:results}.

Overall, we advocate for a conceptual shift in architecture design for conditional generation, which should recenter on caching and make it integral to a model's operation rather than an afterthought. Our contributions are summarized as follows:

\begin{enumerate}
    \item \emph{Cacheability}: We provide evidence that encoder-decoder architectures are good candidates for conditional generation since our cache-friendly models \emph{enhance model performance} compared to ICL, while \emph{reducing cache memory footprint} by more than 98\%.
    
    \item \emph{Parameter efficiency}: We show that \emph{training a few cross-attention layers} (and optionally, a small encoder) suffices to convert decoders into encoder-decoder pairs. We contribute a mix of training tasks that enable context-conditional generation without costly ICL.
    
    \item \emph{Decoder-as-encoder}: We show that representations extracted from pre-trained causal decoders \emph{can be used as-is} to replace an encoder.
\end{enumerate}

\section{Caching Representations}
\label{sec:caching}
Let $\context$, $\query$ and $\answer$ denote sequences of tokens from a shared vocabulary $\mathbb{V}$. We write $\lvert \context \rvert$, $\lvert \query \rvert$ and $\lvert \answer \rvert$ the respective length of these sequences.
Figure \ref{fig:cond_gen_and_caching}(a) illustrates an LLM which, conditioned on $\context$, produces an $\answer$ to a user-specified $\query$.

\paragraph{Assumptions.} In what follows, we make three main assumptions:
(A1) the context is not unique and depends on the query;
(A2) there exists a manageable amount of contexts to condition on; and
(A3) the $\context$ is large, \ie, $\lvert \query \rvert + \lvert \answer \rvert \ll \lvert \context \rvert.$
In doing so, we restrict ourselves to the regimes where processing contexts offline is both compelling (A1 and A3) and viable (A2).

Many ICL use cases satisfy these assumptions.
For example, when facing a
number of different tasks, we may craft task-specific instructions, each detailing how to obtain the $\answer$ to
the $\query$ (Figure~\ref{fig:cond_gen_and_caching}(b)).
The same applies to RAG-based question-answering when retrieved documents are encoded independently: the retriever selects from a corpus the most relevant documents ($\context$) to answer the user question ($\query$).

\subsection{Inference Time Complexity}
\begin{figure}[t]%
\begin{subfigure}[t]{0.45\linewidth}%
\centering%
\includegraphics[width=\textwidth]{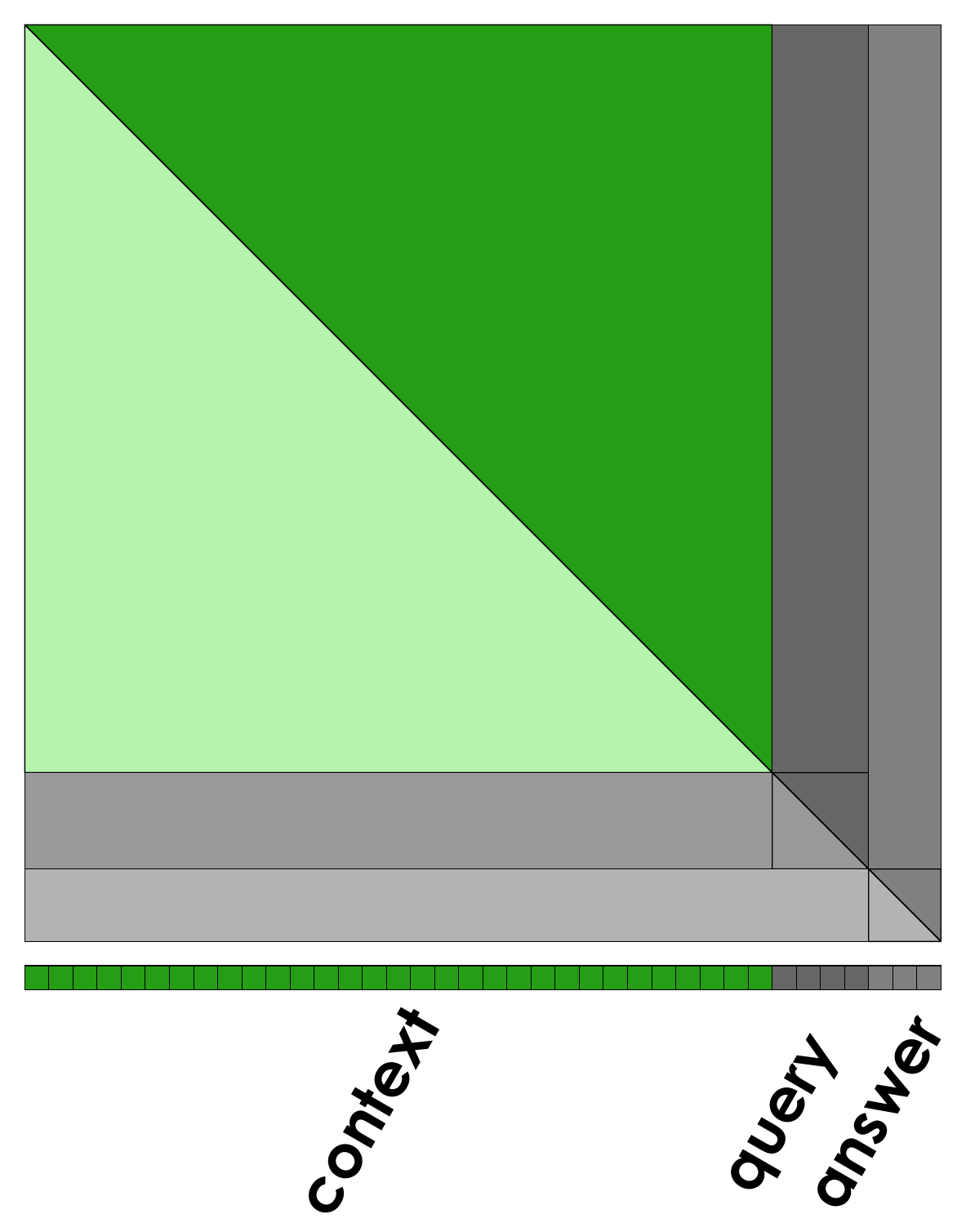}%
\caption{Without cache.}%
\end{subfigure}%
\hfill%
\begin{subfigure}[t]{0.45\linewidth}%
\centering%
\includegraphics[width=\textwidth]{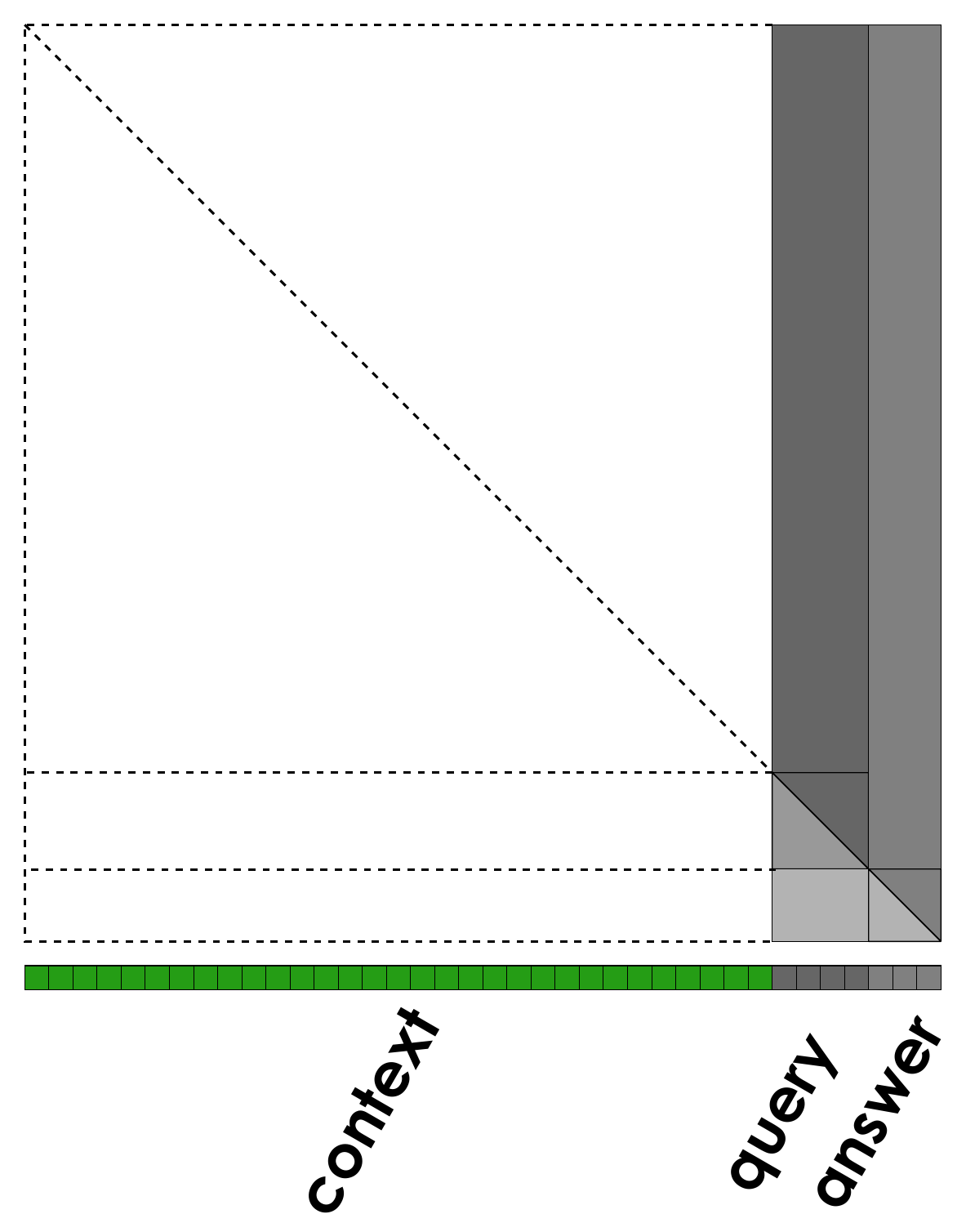}%
\caption{\kvcache{} caching.}%
\end{subfigure}%
\caption{%
\textbf{Stylized representation of attention execution time} (area) for context-conditional language modeling.
Dashed lines in (b) shows the savings when the $\context$'s (past) keys and values $\cache$ are provided.
For causal models, the area below the diagonal represents execution time that could be saved by other means.
}%
\label{fig:kv-cache}%
\end{figure}

In general, the time complexity of LLM inference is dominated by its attention mechanism, which in the ICL setting is $\Ohfull$.
In the large context regime (A3), this simplifies to $\Ohquad$: we can thus expect tangible inference speedups by improving how we handle the $\context$.
One way to achieve such speedups is to pre-process the contexts offline to some intermediate states ($\cache$) and provide it to the model at inference (Figure~\ref{fig:cond_gen_and_caching}(c)).
This way, the quadratic cost of processing the context is paid once, allowing inference to simply look up a ready-made $\cache$ (Fig.~\ref{fig:cond_gen_and_caching}(d)).
Note that the incurred storage and communication overheads are linear in the size of the $\cache$, which is linear in $\context$ length.
In this setup, inference time complexity becomes $\Ohcacheinfer$, \ie{} linear in $\context$ length (Fig.~\ref{fig:kv-cache}), a significant speedup.

\subsection{Practical costs of caching}

Practical considerations might negatively affect a cached-enabled setup.
\textbf{Loading} and \textbf{communication} overheads are both linear in $\cache$ size, motivating smaller $\cache$.
Any \textbf{extra operations} required at inference mitigates caching benefits.
Caching methods may incur an (implicit or explicit) cost in the \textbf{quality} of the generated $\answer$.
See Appendix~\ref{sec:app-caching} for details.

\begin{figure*}[t]
    \centering
    \begin{subfigure}[b]{0.45\textwidth}
        \includegraphics[width=\linewidth]{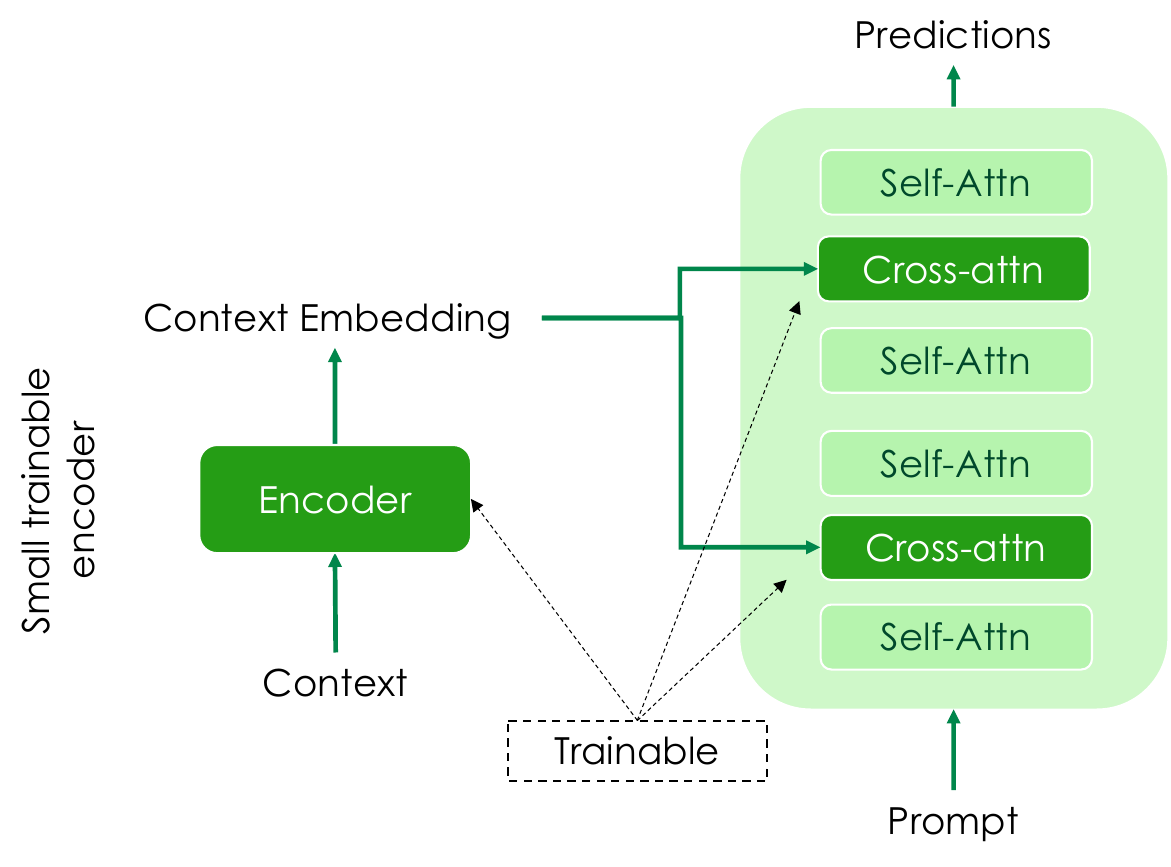}
        \caption{Trainable small bi-directional encoder.}
        \label{fig:model_architecture_encoder}
    \end{subfigure}
    \hfill
    \begin{subfigure}[b]{0.45\textwidth}
        \includegraphics[width=\linewidth]{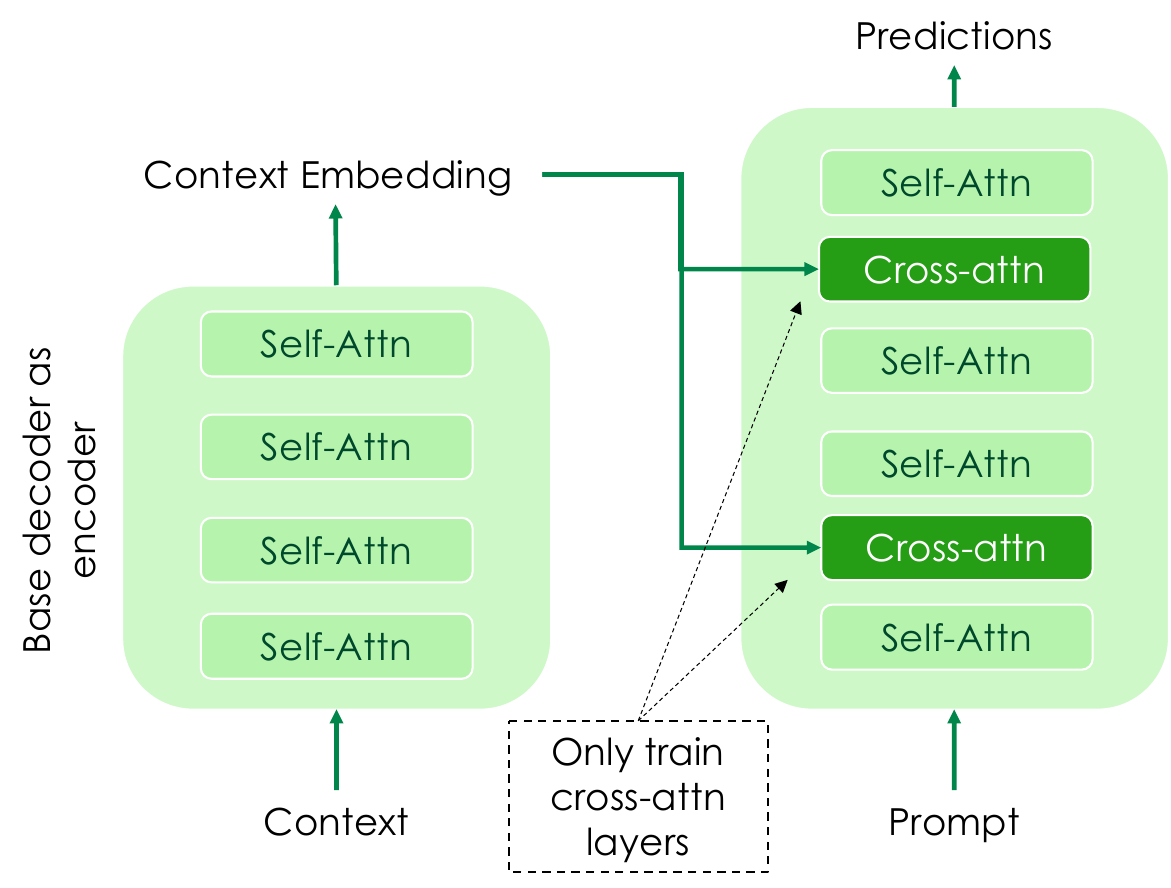}
        \caption{Decoder as the (causal) encoder.}
        \label{fig:model_architecture_decoder_as_ecoder}
    \end{subfigure}
    \caption{\textbf{\ourmodel{}'s architectures.} A decoder-only model implements encoder-decoder architectures. Fine-tuning out in a parameter-efficient fashion via training only a small number of cross-attention layers.\label{fig:model_architecture}}
\end{figure*}

\subsection{Approaches to Caching}

\paragraph{\kvcache{} Caching.} The default approach 
is to store the (past) key and value states generated while processing $\context$, hereafter called \kvcache{} caching.
\kvcache{} caching is commonly associated with a conversational setting, in which case the $\cache$ often remains on the GPU device between conversational rounds (Figure~\ref{fig:cond_gen_and_caching}(d)).
Other setups, such as serving multiple conversations, demand however that we move $\cache$ from storage (and/or CPU RAM) to the GPU, incurring further latency.
As an example, for \llama{}-7B~\cite{touvron2023llama} using 16~bits precision, we must move around a whopping 512~kB \emph{per token}.\footnote{%
32 layers, 32 heads per layer, 128 floating points per key \emph{and} value (thus a factor 2), 2 bytes per 16 bit floating point}
Smaller per-token $\cache$ sizes are thus desirable.

\paragraph{\jitkvcache{} Caching.} An alternative approach is to 
store the (past) hidden states of the model in the $\cache$
(in the case of \llama{}-7B, this would be half as big, \ie{}, 256~kB per token).
At inference time, once these hidden states are loaded on GPU, 
we can recover the full keys and values in $\Ohlinear$ well-parallelized operations (recall that a transformer's keys and values are obtained by a linear operation on the hidden states).
We call this approach ``just-in-time'' or \jitkvcache{} caching.

\paragraph{\xc{} Caching (Ours).}
Both \kvcache{} and \jitkvcache{} caching perform the exact same computations as the original model without a cache. 
They both suffer from two types of costs while producing the same $\answer$:
the \textbf{size} of the $\cache$ and the \textbf{operations} to be performed at inference (\eg, convert hidden states to keys and values).
In this work, we propose \xc{} caching (read \emph{cross-context caching}), which considers the \emph{quality loss} of the generated answer as a third such cost and present two models balancing these three costs. 
Both models employ cross-attention layers integrated into the decoder architecture, enabling the model to attend on context cached with a compact memory footprint at inference time. 
We describe both models in the next section. 

\section{\xc: Cross Attending to Efficiently Cached Context}
\label{sec:x_models}

To reduce the memory footprint of caching, our approach draws inspiration from the encoder-decoder architecture, which until recently was the go-to design for conditional generation.
Recently, decoder-only models have become more popular in part due to their data efficiency: the entire parameter budget is allocated to the decoder and not ``wasted'' on an encoder, with the additional advantage that all parameters are trained against all data tokens.
Even though maintaining an external encoder may seem wasteful, we favor an encoder-decoder architecture: it better lends itself to pre-computing and caching $\context$ representations.
Indeed, only the encoder output vectors need to be stored, as opposed to intermediate states across all of the decoder's self-attention layers (\kvcache{} and \jitkvcache{} cache).

In what follows, we refer to a model as composed of an encoder $\mathcal{E}:\mathbb{V}^{\lvert \scriptcontext \rvert} \mapsto \mathbb{R}^{d\lvert \scriptcontext \rvert}$, which takes in a $\context$ and outputs token-level representations of size $d$, and of a decoder $\mathcal{D}:\mathbb{V}^m \times \mathbb{R}^{d\lvert \scriptcontext \rvert} \mapsto \Delta^{|\mathbb{V}|}$, which takes as input the query and the context encodings and outputs an answer in the simplex of size $|\mathbb{V}|$.
More precisely, the decoder $\mathcal{D}$ autoregressively outputs the parameters of a categorical distribution over the vocabulary.

For the sake of parameter efficiency and to leverage state-of-the-art pre-trained LLMs, we start from an existing decoder-only model and augment it with new cross-attention layers interleaved between existing self-attention layers, as illustrated in Figure~\ref{fig:model_architecture}. We consider two strategies to implement the encoder $\mathcal{E}$. 
The first one trains a \emph{small encoder}, \ie{}, $\mathcal{E}$ is orders of magnitude smaller than $\mathcal{D}$. 
The second one uses a \emph{decoder as encoder}, \ie{}, the frozen decoder-only model is used out-of-the-box as the encoder ($\mathcal{E} := \mathcal{D}$).
More precisely, we use as encodings the representations extracted from the pre-trained~$\mathcal{D}$ at its last layer before the language modeling head. 

Choosing one approach over the other depends on practical considerations.
If caching is possible and context representations can be computed offline, then using the decoder as an encoder is preferable for overall simplicity and parameter efficiency. 
Otherwise, a small encoder would make just-in-time processing of contexts significantly less costly.
Both approaches inherit the advantages of trained decoder-only models while benefiting from using an encoder during inference.
In particular, contextual information can be efficiently cached since only the output at $\mathcal{E}$'s top layer must be stored instead of the entire set of $\mathcal{D}$'s intermediate states. 

Finally, to enable context conditioning, we train exclusively the newly added modules: the cross-attention layers in both settings and the small encoder in the setting that requires it.
The base decoder is kept frozen in both settings (even when it acts as an encoder).
Consequently, our training procedure does not affect the original parameters of $\mathcal{D}$, which can still be used as a general-purpose language model when the additional layers are removed. 

\section{Experimental Setup}
We focus on the question-answering task and train on a combination of datasets where $\context$, $\query$, and $\answer$ triplets are available. 
QA is an ideal testbed for our approaches as it requires efficiently looking up external information and accounting for it at generation time. In other words, QA almost isolates the ability of the model to “read” the context, which is what we needed to test for.

\subsection{Training Dataset}
\label{sec:training_dataset}

We build a training dataset by standardizing and pooling together the training partitions of five publicly available and diverse (open-domain, multi-hop, conversational) datasets: \textsc{Natural Questions} (NQ)~\citep{kwiatkowski2019natural}, \textsc{HotpotQA}~\citep{yang2018hotpotqa}, \textsc{TopiOCQA}~\citep{adlakha2022topiocqa}, 
\textsc{MS MARCO}~\citep{bajaj2016ms}, and \textsc{Squad-V2}~\citep{rajpurkar2018know}. Examples from the resulting training dataset contain a $\query$ (natural-language question), an $\answer$ (expected output), and one or more $\context$s (\eg{}, knowledge base articles), where at least one of them contains the answer to the query. We refer to this context as \emph{reference} $\context$. 
We make use of a validation set for model selection and hyperparameter tuning. For datasets with a validation set but not a test partition, we use the validation set for testing and create a validation set by holding out 10\% of randomly selected training samples.
We apply a further filtering step in our training data and remove examples with contexts longer than 6,000 tokens, corresponding to less than 5\% of the samples.

Evaluation is performed on the resulting test partitions of \textsc{Natural Questions}, \textsc{HotpotQA}, and \textsc{TopiOCQA}. Further details and statistics of the datasets we considered can be found in Appendix~\ref{sec:app-datasets}.

\subsection{Auxiliary Tasks}
\label{sec:dssk_training}

In addition to training on the primary QA tasks,
we optimize our models on context repetition tasks, as described below.
The advantage of defining such auxiliary tasks is two-fold.
On the one hand, they allow us to optimize the likelihood of all available tokens, even those used as input to the encoder.
On the other hand, they help avoid sub-optimal solutions where the cross-attention layers behave as simple identity operators. 
Indeed, our models could learn to ignore the context but cannot do so when tasked to repeat the context.

\begin{figure}[t]
    \centering
    \includegraphics[width=0.7\linewidth]{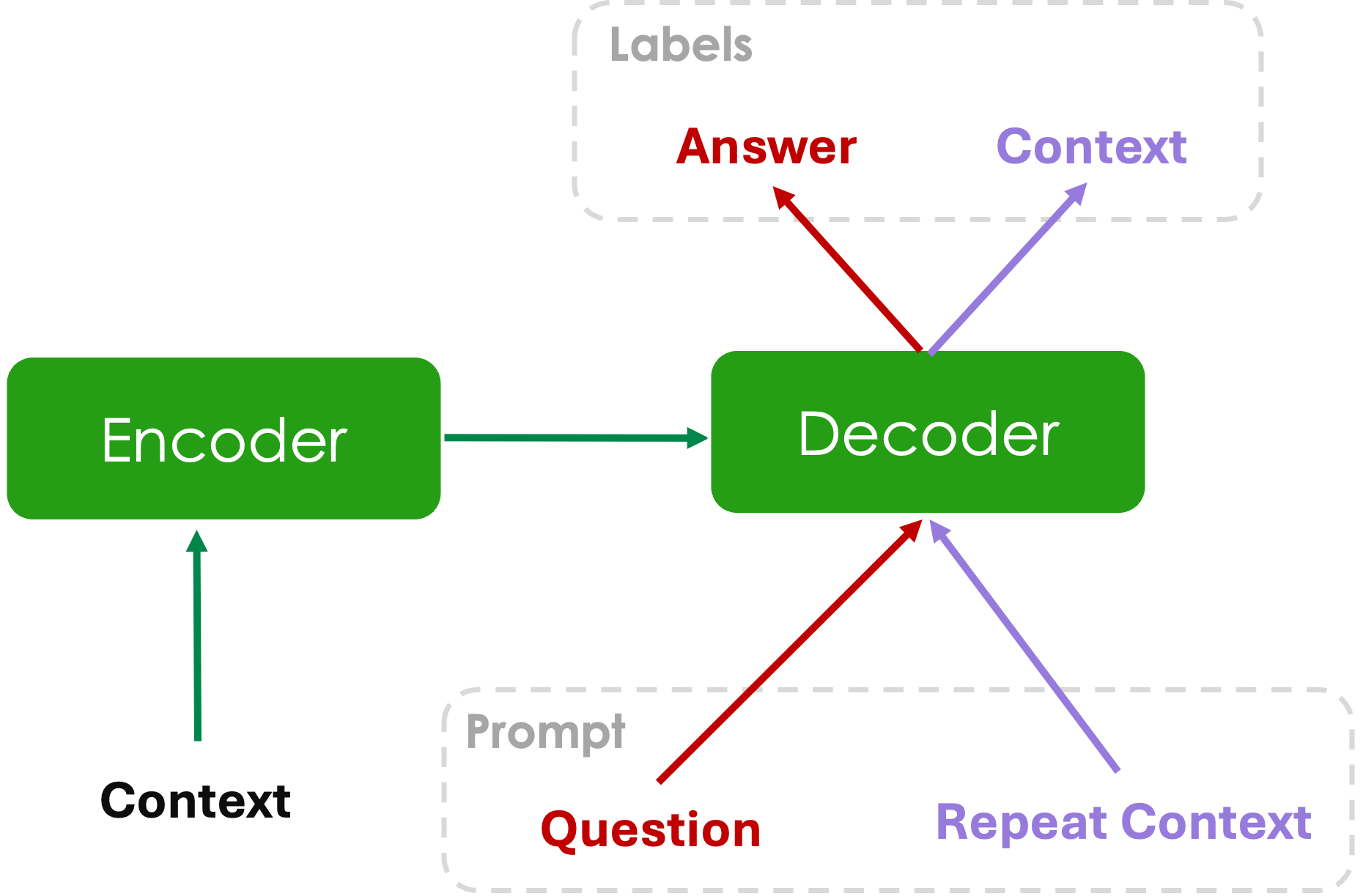}
    \caption{\textbf{Multitask training strategy.} Within an epoch, each example is presented to the model twice. In the first round, the model predicts the answer conditionally based on the context and the query. In the second appearance of an example within the epoch, we train the model to repeat or infill the context.}
    \label{fig:dssk_training}
\end{figure}

In practice, we use reserved tokens to instruct the model to repeat the context, either as-is or by infilling~\citep{bavarian2022efficient}, \ie{} returning it in the prefix-suffix-middle or the suffix-prefix-middle order, as done by~\citet{li2023starcoder,lozhkov2024starcoder}.
Such tasks introduce new variability, as the model learns to copy-paste the context and to find and replace its missing chunks, resulting in improved performance in the multi-epoch training setting. 
Note that we train our models on question-answering and context repetition on every training sample. We set one training epoch to correspond to two passes over the training dataset, where we perform the primary or the secondary tasks in each pass, as illustrated in Figure~\ref{fig:dssk_training}.

\subsection{Implementation Details}
\label{sec:models_finetunning_details}

We rely on the openly available pre-trained \llama{}~\citep{touvron2023llama} to define variations of \ourmodel{}. Specifically, all our empirical assessments use the 7 billion parameter version of \llama{}. For the variation of \ourmodel{} where a dedicated encoder is trained (referred to as \ourmodelencoder{} from now on), we fine-tune as encoder a \textsc{LongFormer}~\citep{beltagy2020longformer}, which is a \textsc{BERT}~\citep{devlin2018bert} of approximately 125M parameters,  pre-trained on relatively long documents using efficient self-attention.
We note, however, that the longest publicly available \textsc{LongFormer} accepts inputs of at most 4,096 tokens, which is shorter than our maximum length of 6,000. We then increase the model's maximum input length by repeating the starting position embeddings. Upon fine-tuning, the model does handle contexts longer than 4,096 tokens without noticeable problems.

As per adding cross-attention layers to \ourmodel{}, we introduce one cross-attention layer every few self-attention layers of the transformer. In particular, we found the 5-6 configuration (\ie{}, inserting five cross-attention layers while skipping six self-attention ones) to work consistently well. We thus use this configuration throughout our evaluations. We remark that we strive to be parameter efficient and keep the parameter count of added modules below 10\% of $\mathcal{D}$'s parameter count.
Training is carried out with the \textsc{AdamW} optimizer~\cite{loshchilov2017decoupled} with a batch size of 256 and for 40,000 steps (amounting to 4 training epochs). 
We use a linear learning rate scheduler with a warm-up phase and such that the learning rate reaches 0 at the last training step. A comprehensive list of hyperparameter values is shown in Table~\ref{tab:hyperparams} in Appendix~\ref{sec:app-exp}.

\subsection{Metrics and Evaluation}

We compare our models against ICL methods for generating answers conditioned on $\context$ and $\query$. We note that contexts in the cases we considered present a relatively low \emph{signal-to-noise} ratio, as most of the tokens are related to the answer but are not at all relevant.
In some more extreme situations, the posed question cannot be answered from the context, and models are expected (and trained or prompted to) indicate that it is not possible to answer based on the provided context.

We use the same metrics and evaluation setup described by~\citet{adlakha2023evaluating} -- such as \textsc{F1 score}, \textsc{recall}, \textsc{Meteor}, and \textsc{RougeL} -- but keep \textsc{F1} as our metric of focus. In addition, we evaluate \textsc{BERTScore}~\citep{zhang2019bertscore} measured between predictions and ground-truth answers.

\section{Results}
\label{sec:results}

\subsection{Comparison with existing methods}
We first compare our method to existing approaches for conditional generation.
Our main baseline is ICL, \ie{}, providing the context as part of the prompt. 
More specifically, we report baseline results for \llamachat{}, which we found to perform better than the base pre-trained \llama{}. We further report results for \textsc{GPT-3.5-Turbo}. 
For these ICL baselines, we selected the prompt templates based on generated answer quality on sample validation data (refer to Appendix~\ref{sec:app-exp} for details).
Finally, we report the results of \textsc{Fusion-in-Decoder} (FiD)~\citep{izacard-grave-2021-leveraging}, a \textsc{T5}-based~\citep{raffel2020exploring} conditional generative model, which consistently proved to be state-of-the-art on QA tasks~\citep{borgeaud2022improving,wang2023shall,adlakha2023evaluating}. 
Unlike the decoder-only backbone of our models, FiD is arguably no longer a general-purpose model, as 
all of its parameters are fine-tuned to perform QA. More importantly, as discussed in more depth in Section~\ref{sec:broader_analysis}, \textbf{pre-processing and caching context representations is not an option for FiD since it requires knowing the question at encoding time}.
Nevertheless, we introduce FiD as a reference, to check where our models stand relative to established QA-specialized models.

Results are presented in Table~\ref{tab:results_icl}. 
On the considered datasets, cross-attending to the contexts (\ourmodel{} or \ourmodelencoder{}) considerably improves performance w.r.t prompting (\llamachat). 
The gap varies depending on the dataset. We conjecture that this is due to the high variance induced by prompting, although there might exist an optimal prompt for each dataset to help close this gap. Thus, approaches that do not rely on the prompt offer the advantage of being more broadly applicable and, hence, more practical. We also note that even in the setting where the decoder is used as an encoder, cross-attending to contexts still yields better performance than ICL, no matter the base decoder we compare against. This suggests that the trained cross-attention layers compensate for the potential sub-optimality because the encoder representations are not explicitly trained for the task we evaluate. A broader set of results comprising more metrics and models can be found in Table~\ref{tab:correctness_main_results}.

\setlength{\tabcolsep}{4pt}
\begin{table*}[t!]
  \centering
  \small
  \begin{tabular}{llccc}
\toprule
Dataset           & Model                     & \textsc{Cache size} ($\downarrow$)       & \multicolumn{1}{l}{\textsc{F1} ($\uparrow$)} & \multicolumn{1}{l}{\textsc{BERTScore} ($\uparrow$)} \\ \midrule
                  & \textsc{GPT-3.5 Turbo}    & Unknown                        & 57.80                           & 90.87                                  \\
                  & \textsc{FiD}              & Non-cacheable                        & 59.05                           & 91.75                                  \\ \cmidrule{2-5} 
\textsc{NQ}       & \llamachat                & 512\phantom{.0}           & 41.26                           & 87.43                                  \\
                  & \ourmodel{} (Ours)        & \phantom{01}8\phantom{.0} & 59.95                           & 92.87                                  \\
                  & \ourmodelencoder{} (Ours) & \phantom{01}\textbf{1.5}  & \textbf{63.12}                  & \textbf{93.30}                         \\ \midrule
                  & \textsc{GPT-3.5 Turbo}    & Unknown                        & 39.37                           & 87.83                                  \\
                  & \textsc{FiD}              & Non-cacheable                        & 45.60                           & 90.56                                  \\ \cmidrule{2-5} 
\textsc{HotpotQA} & \llamachat                & 512\phantom{.0}           & 29.63                           & 86.02                                  \\
                  & \ourmodel{} (Ours)        & \phantom{01}8\phantom{.0} & 43.94                           & 90.55                                  \\
                  & \ourmodelencoder{} (Ours) & \phantom{01}\textbf{1.5}  & \textbf{54.57}                  & \textbf{92.08}                         \\ \midrule
                  & \textsc{GPT-3.5 Turbo}    & Unknown                        & 40.18                           & 87.52                                  \\
                  & \textsc{FiD}              & Non-cacheable                        & 31.22                           & 85.95                                  \\ \cmidrule{2-5} 
\textsc{TopiOCQA} & \llamachat                & 512\phantom{.0}           & 33.45                           & 86.33                                  \\
                  & \ourmodel{} (Ours)        & \phantom{01}8\phantom{.0} & 45.47                           & 89.16                                  \\
                  & \ourmodelencoder{} (Ours) & \phantom{01}\textbf{1.5}  & \textbf{47.73}                  & \textbf{89.40}                         \\ \bottomrule
\end{tabular}
\caption{Cache memory footprint (kB/token) and Question-Answer performance on three diverse information-seeking tasks. \textsc{GPT-3.5 Turbo} and \textsc{Llama2-Chat} are given reference context through ICL (prompting), while our approach uses cross-attention layers to ingest reference embeddings. The encoder-decoder approach of \ourmodel{} allows for huge savings as it requires storing only the last hidden states instead of \kvcache{} states throughout layers. Cache sizes here assume 16-bit precision.}

  \label{tab:results_icl}
\end{table*}

\subsection{Trading off accuracy against efficiency}
\label{sec:broader_analysis}

The results in the previous section show that adding and fine-tuning dedicated parameters for context-conditioning improves performance relative to prompting. Based on this observation, in this section, we expand our evaluation to consider alternative approaches that leverage a small number of additional parameters that enable models to condition generation on reference contexts. 
In addition to prediction performance, we now also focus on \emph{computational efficiency};
namely, we assess how amenable different models are to pre-processing and caching representations of contexts.

We thus fine-tune \textsc{LoRA} adapters~\citep{hu2021lora} applied to the same \llama{} decoder we use for \ourmodel{}. In other words, we fine-tune our ICL baseline from Section~\ref{sec:results} to control for the effect of prompting in the model performance.
As observed in Table~\ref{tab:results_lora}, such finetuning drastically improves QA accuracy relative to the ICL baselines reported in Table~\ref{tab:results_icl}.
However, despite enabling improvements in prediction performance, relying on \textsc{LoRA}-style model adaptation still necessitates storing all \kvcache{} states throughout every layer to cache contextual information, incurring significant costs. In contrast, models with an encoder require caching the hidden states of only their last layer. In other words, \ourmodel{} variants greatly reduce the caching footprint simply because they require storing only the last hidden states of the encoder. \ourmodelencoder{} reduces space requirements even further, as the representation of \textsc{LongFormer} are lower dimensional relative to that of \llama{}.

Reducing cache size carries substantial practical implications, notably in reducing the memory footprint of a pre-processed corpus by orders of magnitude. This is particularly significant when storing pre-processed representations of vast datasets like the entire Wikipedia. Additionally, reducing the cache size 
leads to runtime savings by mitigating communication costs, as the volume of information transferred from disk to device is markedly reduced. Finally, scaling down the cache size frees up device memory at inference time, enabling longer generation or larger batch sizes for batched inference. 
See Appendix~\ref{sec:app-caching} for details.

We provide a joint view of those two performance components -- prediction and compute -- in Figure~\ref{fig:mem_bertscore}, where we plot aggregate \textsc{F1} and \textsc{BERTScore} across datasets versus the amount of cache per context token required by models. Note that models closer to the top-right corner are preferred since they are highly accurate at a low caching cost. While no method is optimal for both criteria, the Pareto set consists of:
on one end, ICL models fine-tuned with \textsc{LoRA} which offer slightly higher aggregated \textsc{BERTScore} but require substantial caching space;
on the other end, models with an encoder which make small sacrifices in prediction accuracy while significantly reducing the memory footprint. We also note that there is a gap between \ourmodel{} and \ourmodelencoder{}, and the additional parameters introduced by \ourmodelencoder{} yield a boost in accuracy and improve space efficiency. Note that we consider an extra variant of ICL as \textsc{Llama 2-ICL-JIT-KV}, since it performs just-in-time \kvcache{} projection of cached hidden states, trading time for space. 

\setlength{\tabcolsep}{4pt}
\begin{table*}[t!]
 \centering
 \small
 \begin{tabular}{llccc}
\toprule
Dataset           & Model                     & \textsc{Cache size} ($\downarrow$) & \textsc{F1} ($\uparrow$) & \textsc{BERTScore} ($\uparrow$) \\ \midrule
                  & \textsc{GPT-3.5 Turbo}    & Unknown                            & 57.80                    & 90.87                           \\
                  & \textsc{FiD} (fine-tuned) & Non-cacheable                      & \textbf{72.67}           & \textbf{94.76}                  \\ \cmidrule{2-5} 
\textsc{NQ}       & \llama{} (LoRA)           & 512\phantom{.0}                    & 67.38                    & 93.27                           \\
                  & \ourmodel{} (Ours)        & \phantom{01}8\phantom{.0}          & 59.95                    & 92.87                           \\
                  & \ourmodelencoder{} (Ours) & \phantom{01}\textbf{1.5}           & 63.12                    & 93.30                           \\ \midrule
                  & \textsc{GPT-3.5 Turbo}    & Unknown                            & 39.37                    & 87.83                           \\
                  & \textsc{FiD} (fine-tuned) & Non-cacheable                      & 53.54                    & 89.64                           \\ \cmidrule{2-5} 
\textsc{HotpotQA} & \llama{} (LoRA)           & 512\phantom{.0}                    & \textbf{71.97}           & \textbf{94.62}                  \\
                  & \ourmodel{} (Ours)        & \phantom{01}8\phantom{.0}          & 43.94                    & 90.55                           \\
                  & \ourmodelencoder{} (Ours) & \phantom{01}\textbf{1.5}           & 54.57                    & 92.08                           \\ \midrule
                  & \textsc{GPT-3.5 Turbo}    & Unknown                            & 40.18                    & 87.52                           \\
                  & \textsc{FiD} (fine-tuned) & Non-cacheable                      & 41.52                    & 86.54                           \\ \cmidrule{2-5} 
\textsc{TopiOCQA} & \llama{} (LoRA)           & 512\phantom{.0}                    & \textbf{55.41}           & \textbf{90.80}                  \\
                  & \ourmodel{} (Ours)        & \phantom{01}8\phantom{.0}          & 45.47                    & 89.16                           \\
                  & \ourmodelencoder{} (Ours) & \phantom{01}\textbf{1.5}           & 47.73                    & 89.40                           \\ \bottomrule
\end{tabular}

 \caption{Cache memory footprint (kB/token) and Question-Answer performance with ICL, carried out with a model specialized to a prompt template via \textsc{LoRA}. Note that the FiD model was pre-trained on \nq{}, hence its high performance on that dataset. The encoder-decoder approach of \ourmodel{} allows for huge savings as it requires storing only the last hidden states instead of \kvcache{} states throughout layers. Cache sizes here assume 16-bit precision.}

 \label{tab:results_lora}
 \end{table*}

Detailed QA results are reported in Table~\ref{tab:results_lora}. As previously mentioned, our models incur a slight reduction in prediction accuracy but achieve significant space savings, which proves advantageous in various practical scenarios. For a comprehensive overview, please refer to the full results in Table~\ref{tab:correctness_main_results}.

\subsection{Question-dependent context encoding}

Among the baselines we consider in this work, those relying on decoder-only architectures for language modeling hold a fundamental advantage over encoder-decoder models: they have \emph{access to the question} while computing context representations, allowing the generated representations to be tailored to the specific question.
Indeed, in non-caching popular encoder-decoder settings such as FiD~\citep{izacard-grave-2021-leveraging}, contexts are fed into the encoder together with the questions. 

To test the impact of having access to the question while encoding the context on the performance of our model, we performed experiments with a variation of \ourmodelencoder{} where the question is prepended to the inputs of the encoder. 
Table~\ref{tab:question_in_context} summarizes the results of these experiments performed on \nq{} where the column named \emph{Question-in-context} indicates whether the encoder had access to the question while generating context representations.
We observe a performance improvement, nearly closing the gap with the large cache LoRA baseline. 
This observation suggests that question-dependent context representations could be integrated into cacheable encoder-decoder architectures, offering the best of both worlds: lightweight context caching and high prediction accuracy.

\setlength{\tabcolsep}{4pt}
\begin{table*}[ht]
\centering
\begin{tabular}{cccc}
\toprule
Model                     & Question-in-context & \textsc{F1} & \textsc{BERTScore} \\ \midrule
\llama{} (LoRA)           & \cmark             & 67.38       & 93.27              \\
\ourmodelencoder{} (Ours) & \xmark             & 63.12       & 93.30              \\
\ourmodelencoder{} (Ours) & \cmark             & 66.86       & 93.43              \\ \bottomrule
\end{tabular}
\caption{Question-Answer performance on the \textsc{Natural Questions} dataset. 
Including the question, in addition to the context, as inputs to the decoder significantly enhances the QA accuracy of encoder-decoder models.
}

 \label{tab:question_in_context}
\end{table*}

\section{Related Work} 
\label{sec:relatedwork}

\paragraph{Decoders as encoders.} Repurposing pre-trained decoders is becoming a popular approach to leverage powerful language models for applications other than generative modeling. For example, \textsc{GRIT}~\citep{muennighoff2024generative} converts a pre-trained causal decoder into a bi-directional encoder, yielding sentence-level embeddings while maintaining its ability to perform autoregressive generation of text. However, unlike the models we consider, this conversion requires fine-tuning all model parameters instead of additional ones. Parameter-efficient approaches to turn decoders into encoders were also proposed, such as in~\citep{SFRAIResearch2024} and~\citep{behnamghader2024llm2vec}, where a pre-trained \textsc{Mistral} decoder~\citep{jiang2023mistral} is fine-tuned in a contrastive setting using \textsc{LoRA} adapters to yield sentence level representations for retrieval. Closer to our method is \textsc{CodeT5+}~\citep{wang2023codet5+}, which also defines encoder-decoder architectures rather than turning decoders into sentence encoders. Similar to one of our variants (\ourmodel{}), it is assembled from two pre-trained decoders, one used as an encoder and the other as a decoder and linked with a few cross-attention operations. However, \textsc{CodeT5+} requires fine-tuning the entire and relatively large encoder. We show that a good pre-trained decoder has good enough representations, but one can improve upon it efficiently using just a very small trainable encoder.

\paragraph{Conditioning without prompts.} One recent line of work has focused on controlling a model's generation by intervening in its parameters directly~\citep{wang2023rome,zhang2023counterfactual, wang2023retrievalaugmented}, in particular, to either introduce or erase knowledge post-training. Such approaches would typically require accessing and rewriting the internal parameters of a pre-trained language model and are not amenable to frequently changing contexts such as in a Question-Answer setting.

\paragraph{Efficient inference.} There exist several methods for improving inference speed and memory footprint. One approach is to lower the numerical precision or quantize the model weights and/or data, which has been shown to preserve the model accuracy with only 8 bits per weight~\citep{dettmers2022gpt3}, or even lower precision with 4, 3, or even 2 bits per weight~\citep{frantar2023gptq}. Alternatively, the key-query cache can be compressed~\citep{ainslie2023gqa,nawrot2024dynamic}, although doing so requires fine-tuning. Moreover, methods that compress the cache along the time axis have been proposed~\citep{bulatov2022recurrent,ge2024incontext,chevalier-etal-2023-adapting}. However, compression rates for those cases are not really comparable to having a separate encoder and cross-attending to its outputs. For instance, \citet{ge2024incontext} report a compression rate of at most 4 (as opposed to at least 32 as we claim), and that comes at a cost since compression along the time dimension potentially discards relevant content, as discussed by \citet{chevalier-etal-2023-adapting}. With an architecture similar to ours, \cite{yen2024long} report benefits in having a separate encoder. Although we show that applying cross-attention operators after just a few self-attention layers suffices, and that representations of a pre-trained decoder perform reasonably well without further training. Finally, using Flash Attention~\citep{dao2022flashattention} leads to significant savings for just-in-time processing of contexts. These methods orthogonally complement what we present in this paper and can be combined with \ourmodel{}.

\section{Conclusion} 
\label{sec:conclusion}

We introduced \ourmodel{} as an approach to transform a pre-trained decoder-only language model into an encoder-decoder architecture that can condition generation on both the encoder inputs and the decoder query. This is achieved by integrating cross-attention layers interleaved in between existing self-attention layers of the pre-trained decoder. We describe two approaches for defining the encoder: using a copy of the decoder or introducing a trainable but small bi-directional encoder.

The proposed architecture allows for a reduction in caching space by a factor exceeding 300. When evaluated in the QA setting, we observe a higher prediction accuracy than what standard ICL approaches achieve, either with \llama{} or \textsc{GPT-3.5 Turbo}. Additionally, we observe accuracy levels nearly on par with caching-intensive fine-tuned prompted models, providing a more caching-friendly alternative to prompted language models that proves highly practical. We note that the encoder-decoder strategy for handling contexts separately offers the extra advantage of freeing prompt space, which is now exclusively used for user interaction. Identified limitations are discussed in Section~\ref{sec:limitations}.

\section{Limitations}
\label{sec:limitations}

Most of the models discussed in this paper have impressive results in the QA setting. However, our experience working with these models reveals limitations 
primarily stemming from their reliance on an underlying language model. As such, these methods inherit potential flaws of the language model they build upon.

For instance, typical large language models are trained on vast amounts of text, likely including information related to questions in publicly available QA benchmarks. While such training on related data may offer a shortcut to models for correctly answering questions in the context that they have ``memorized'' during training, it can also introduce errors: the model may ``remember'' related but inaccurate information relative to a specific query. In simpler terms, prompting-based ICL approaches and potentially our models can bypass context and rely solely on memory to generate a continuation. This is undesirable as we aim to ensure that models accurately account for the reference context rather than pre-training's. Fine-tuning models address the abovementioned issue by encouraging reliance on the provided context, as we explicitly do with context repetition and filling auxiliary tasks. However, it also specializes in fine-tuning data to such an extent that it may hurt the generalization ability of datasets that deviate from what was observed during fine-tuning. We estimate to what extent models know the answer after pre-training in Table~\ref{tab:correctness_noICL}.

\paragraph{Unseen Datasets.}
To test the generalization ability of the models on out-of-distribution data, non-identically distributed with respect to fine-tuning one, we used a pre-release version of RepLiQA~\cite{monteiro2024repliqa}.
This is a curated test dataset containing 16,290 reference documents, with roughly 5 question-answer pairs per document.
The documents were written by human annotators, who wrote about imaginary scenarios, each featuring several subsections with titles and subtitles.
Importantly, given that the reference documents in this dataset contain fictional or untrue information, we can reasonably assume that none of the models can rely on or be misled by their memory of the training data. Additionally, samples in this dataset exhibit unique styles that may differ from the fine-tuning data.

\begin{figure}[t]
    \centering
    \includegraphics[width=\linewidth]{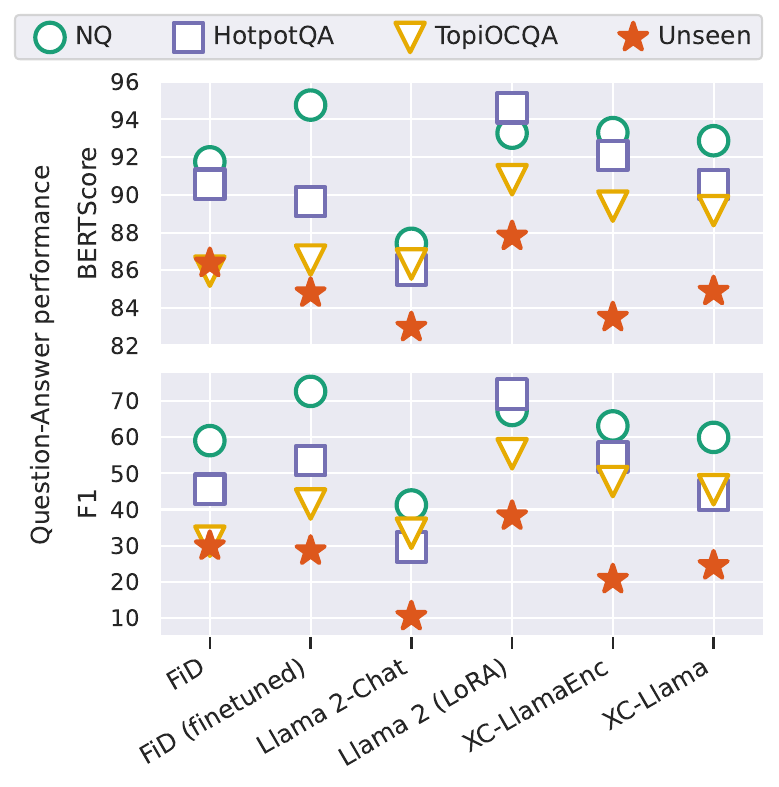}
    \caption{QA performance of various models on our three test datasets against our curated unseen dataset. All models struggle to generalize in this out-of-distribution scenario where the test dataset significantly differs from that used for training.}
    \label{fig:repliqa_results}
\end{figure}

When evaluated in this more challenging out-of-distribution scenario, the accuracy of all models experiences a significant decline, as reported in Figure~\ref{fig:repliqa_results}, where we compare \QA{} scores for all models across all datasets, including the pre-release version of RepLiQA labeled as \emph{Unseen}.
We hypothesize that, apart from the inability of models to rely on memory to answer queries, the main factor driving the accuracy decline is the divergence between the test dataset and the fine-tuning data. This discrepancy may arise from variations in writing style, document lengths, signal-to-noise ratio, and the presence of distracting content related to the question but not useful for its answer, among others. Identifying the exact sources of errors and enhancing the robustness of context-conditional models are promising avenues for future research.

\section*{Ethics Statement}
We acknowledge the environmental impact of computational resources, including the carbon footprint associated with data storage and processing. By focusing on conditional generation through caching with a small memory footprint, we aim to minimize energy consumption and promote sustainability in computational practices. 

While cache reduction can lead to speed improvements, we recognize the importance of additionally implementing safeguards to ensure responsible usage of computational models.  This entails mitigating bias, promoting fairness, and preserving privacy, thus fostering ethical deployment of technology.

\section*{Acknowledgements}
We would like to thank Christian Hudon for technical support and enabling scaling up our experiments. We also thank Siva Reddy for insightful discussions that helped shape this work.

\clearpage
\appendix

\onecolumn

\vskip0.5cm

\section{Practical costs of caching}
\label{sec:app-caching}
Although a complexity linear in the context length is achievable while maintaining  $\context$ quality, practical considerations might negatively affect a cached-enabled setup.

\paragraph{Loading and communication overheads.}
Even though long-term storage costs are relatively low, loading and communication latencies are directly proportional to the space footprint of the cached content.
The two main bottlenecks here are loading $\cache$ from disk (if disk storage is used), and moving $\cache$ to the GPU's memory.
If the total amount of information to cache fits in CPU RAM, there may be no need to store them on disk during inference, enabling better-than-linear returns on smaller caching sizes.

\paragraph{Extra operations.}
In the case of \llama{}, \jitkvcache{} caching requires a $\cache$ that is half as big as the one of \kvcache{} -- 254~kB instead of 512~kB per token -- and is thus clearly more advantageous on the loading and communication fronts.
However, extra operations are required at inference time to convert the hidden states into the keys and values required by the model.
Here these extra operations turn out to be cheap linear operations.
Similarly, our \xc{} caching approaches require very little storage (\ie{}, 8~kB and 1.5~kB per token), which is a clear win on loading and communication fronts.
However, extra layers are added to the original model in Sec.~\ref{sec:x_models}), which induces time costs that have to be accounted for in the overall speedup.
Again, these extra operations turn out to be negligible compared to the benefit of caching.

\paragraph{Explicit quality cost.}
\kvcache{} and \jitkvcache{} caching both return \emph{exactly} the same $\answer$ as the corresponding un-cached model: these caching strategies incur no (explicit) cost in $\answer$ quality compared to that un-cached model.
In contrast, our \xc{} caching strategy does not confer this guarantee: compromises between costs and benefits must be considered.

\paragraph{Implicit quality cost.}
All caching methods considered in this work share the same implicit ``opportunity cost'': processing $\context$ offline implies that the resulting $\cache$ is independent of the specifics of the $\query$
Depending on the use case, this might result in a loss of $\answer$ quality.
For instance, in the \QA{} setting, knowing the question when processing the context allows the model to focus on the information that is useful to generate an $\answer$.
Arguably, this results in better context representations as it reduces the risk of confusing the LLM with irrelevant information. 

\vskip0.5cm

\section{Timing cache loading}
\label{sec:app-caching_time}

Table~\ref{tab:cache_loading_time} presents a comparison in loading time of cached context all the way from disk to device memory for each approach. Results correspond to varying sequence lengths -- $2^{\{10, 11, 12, 14\}}$ -- for a batch size of 8, assuming a transformer with 32 layers and hidden dimension of 1024. We used 16-bit precision. Timing was carried out 100 times, and we discarded the initial 10 measurements. Results in the table indicate a 95\% confidence interval. As the results show and one would expect, the reduced cache size due XC-cache reduces loading time by two orders of magnitude. These results are now included in the paper in a new appendix.

\begin{table}[ht]
\resizebox{\textwidth}{!}{
\begin{tabular}{ccccc}
\toprule
\multicolumn{1}{l}{} & \multicolumn{4}{c}{Context length}                                                        \\ \cline{2-5} 
\textit{\textbf{}}   & \emph{1024}                 & \emph{2048}                 & \emph{4096}                 & \emph{8192}        \\ \midrule
\emph{KV-Cache}             & 0.893991 +- 0.004639 & 1.748637 +- 0.008972 & 3.542025 +- 0.010325 & 6.841477 +- 0.008582 \\
\emph{XC-Cache}             & 0.012668 +- 0.000806 & 0.034943 +- 0.000939 & 0.063986 +- 0.000778 & 0.120361 +- 0.001406 \\ \bottomrule
\end{tabular}}
\caption{Cache loading time in seconds of cached context all the way from disk to device memory for each approach. Results correspond to varying sequence lengths -- $2^{\{10, 11, 12, 14\}}$ -- for a batch size of 8, assuming a transformer with 32 layers and hidden dimension of 1024. We used 16-bit precision. Timing was carried out 100 times, and we discarded the initial 10 measurements. Results in the table indicate a 95\% confidence interval.}
 \label{tab:cache_loading_time}
\end{table}
\section{Datasets}
\label{sec:app-datasets}
\begin{wraptable}{r}{6.5cm}\
 \centering
 \small
 \begin{tabular}{lrrr}
     \toprule
     Dataset & \# Train & \# Valid & \# Test\\
     \midrule
     Natural Questions & 83,118  & 2,240   & 5,067 \\
     HotpotQA          & 83,038  & 7,405   & 7,405 \\ 
     TopiOCQA          & 42,702  & 2,656   & 2,512 \\
     MS Marco          & 808,298 & 101,035 & 0     \\
     Squad V2          & 130,293 & 11,872  & 0     \\
     \midrule
     Totals & 1,147,449 & 125,208 & 14,084 \\
     \bottomrule
 \end{tabular}
 \caption{Statistics for datasets used to train and test our models. For the first three datasets, we use the validation split of each to test our model and split the provided train set into \emph{train} and \emph{valid}.}
 \label{tab:dataset_stats}
 \end{wraptable}
In this section we give some details about our training datasets and how we used them for training. 
Statistics about the datasets are in Table~\ref{tab:dataset_stats}.

\paragraph{Natural Questions (\nq{}).} Aimed at evaluating open-domain question-answering tasks, Natural Questions~\cite{kwiatkowski2019natural} presents a collection of real user questions from Google queries. Answers are written by human annotators and are based on Wikipedia articles (that may or may not contain the exact information needed for the question). 
For each sample in the dataset, the context we use as input is a paragraph containing the answer. 
We use the open version of \nq~\cite{lee2019latent}, which is the subset of the original dataset where the contexts have at most 100 tokens from English Wikipedia (indexed Dec 2018). This open variant of \nq{} does not provide the contexts relevant to each question. We thus use the questions in \textsc{open} \nq{} to query the original \nq{} dataset and fetch the contexts associated with each question. 
Given that \nq{} provides multiple correct answers to a given $\query$-$\context$ pair, we split each sample into multiple rows at train time, so that each question-context pair is paired with each one of the provided answers. 
For the test set, however, we evaluate a test sample once and compare model outputs to all the reference $\answer$s and consider the answer that maximizes the given metric.

\paragraph{Hotpot QA.} The dataset contains open-domain questions that require at least two or more documents to be answered~\cite{yang2018hotpotqa}. Each sample in the dataset includes a list of contexts (paragraphs from English Wikipedia indexed Dec 2017), two of which contain useful information for answering the question, while the remainder are distractors. Thus, the answers in this dataset typically depend on more than one component of the context, so a model is expected to look up different parts of the context to compose an accurate answer.

\paragraph{TopiOCQA.}  A conversational dataset where samples represent information-seeking open-domain dialogue, the answer to each question in TopiOCQA~\cite{adlakha2022topiocqa} is based on 200-token passages from English Wikipedia (indexed Oct 2020). Human annotators generate these reference answers.

\paragraph{MS MARCO.} Intended to evaluate reading comprehension
and question answering, the questions from the MS MARCO dataset~\cite{nguyen2016ms} are sampled from real user questions from Bing or Cortana. The reference passages are collected from the web (not necessarily Wikipedia) through the Bing search engine, and they may or may not be enough to answer the question. Answers are human-generated.

\paragraph{Squad V2.} 
Squad v2~\cite{rajpurkar2018know} consists of a small set of question/answer pairs generated by annotators and based on Wikipedia articles. The answers in Squad are spans from the original context or yes-no replies and hence are less abstractive.

\paragraph{}
We train our baselines and models with the train split of all five datasets mentioned above, concatenated and shuffled. 
We evaluate models on the validation split of the first three datasets -- Natural Questions, Hotpot QA, and TopioCQA -- as well as an unseen dataset we created.

\vskip0.5cm

\section{Complete set of results}
\textbf{\label{sec:app-exp}}
Table \ref{tab:correctness_main_results} contains results for all methods, metrics, and for all datasets we accounted for in our empirical assessment.

\renewcommand{\arraystretch}{1.25}
 \begin{table*}[th!]
 \centering
 \small
 \begin{tabular}{llccccccc}
 \toprule
Dataset & Model                     & EM & Precision & Recall & F1 & Rouge-L & METEOR &  \textsc{BERTScore}  \\
         \midrule
\multirow{8}{*}{Natural Questions} 
         & \textsc{GPT-3.5 Turbo}     & 43.08 & 55.10 & 74.87 & 57.80 & 57.67 & 53.22 & 90.87 \\
         & \textsc{FiD}               & 50.64 & 63.18 & 58.63 & 59.05 & 59.02 & 47.06 & 91.75 \\
         & \textsc{FiD} (fine-tuned)  & 67.00 & 75.20 & 72.59 & 72.67 & 72.52 & 58.30 & 94.76 \\
         & \llamachat{}               & 24.96 & 40.41 & 58.20 & 41.26 & 41.28 & 38.04 & 87.43 \\
         & \llama{} (LoRA)            & 59.64 & 70.55 & 67.39 & 67.38 & 67.08 & 53.88 & 93.27 \\
         \cmidrule(ll){2-9}
         & \ourmodelencoder{} (Ours)  & 56.77 & 65.93 & 62.89 & 63.12 & 62.99 & 49.99 & 93.30 \\
         & \ourmodel{} (Ours)         & 51.41 & 62.62 & 60.01 & 59.95 & 60.21 & 47.47 & 92.87 \\ 
 \midrule
 \multirow{8}{*}{HotpotQA} 
         & \textsc{GPT-3.5 Turbo}     & 27.62 & 39.08 & 46.43 & 39.37 & 39.35 & 34.01 & 87.83 \\
         & \textsc{FiD}               & 33.81 & 48.95 & 45.54 & 45.60 & 45.63 & 35.24 & 90.56 \\
         & \textsc{FiD} (fine-tuned)  & 44.04 & 54.49 & 55.38 & 53.54 & 53.42 & 41.78 & 89.64 \\
         & \llamachat{}               & 14.22 & 27.07 & 51.48 & 29.63 & 29.56 & 30.82 & 86.02 \\
         & \llama{} (LoRA)            & 58.33 & 74.91 & 72.72 & 71.97 & 71.91 & 55.27 & 94.62 \\
         \cmidrule(ll){2-9}
         & \ourmodelencoder{} (Ours)  & 43.29 & 56.88 & 54.90 & 54.57 & 54.63 & 41.13 & 92.08 \\
         & \ourmodel{} (Ours)         & 31.90 & 46.27 & 44.02 & 43.94 & 44.35 & 32.17 & 90.55 \\
 \midrule
 \multirow{8}{*}{TopiOCQA} 
         & \textsc{GPT-3.5 Turbo}     & 17.18 & 44.29 & 45.77 & 40.18 & 39.23 & 35.50 & 87.52 \\
         & \textsc{FiD}               & 16.43 & 53.56 & 27.17 & 31.22 & 31.02 & 20.56 & 85.95 \\
         & \textsc{FiD} (fine-tuned)  & 24.94 & 49.80 & 41.69 & 41.52 & 41.31 & 32.31 & 86.54 \\
         & \llamachat{}               & 13.76 & 42.56 & 36.87 & 33.45 & 32.88 & 26.95 & 86.33 \\
         & \llama{} (LoRA)            & 25.89 & 61.29 & 57.43 & 55.41 & 54.92 & 47.19 & 90.80 \\
         \cmidrule(ll){2-9}
         & \ourmodelencoder{} (Ours)  & 24.82 & 55.44 & 47.83 & 47.73 & 47.49 & 39.22 & 89.40 \\
         & \ourmodel{} (Ours)         & 19.17 & 52.49 & 46.64 & 45.47 & 44.94 & 38.37 & 89.16 \\
 \bottomrule
 \end{tabular}

 \caption{Question-Answer performance on three diverse information-seeking tasks. All models in this table that are trained/fine-tuned, did so on the same five datasets; the table reports the metrics of testing these models on different test splits.}

 \label{tab:correctness_main_results}
 \end{table*}

\vskip0.5cm

\section{Inference results with no context }

We evaluate our pretrained LLMs on the test split of our datasets without passing a context to base the answer on. We do this to have a better sense of how many of the answers they already know without ICL.

\renewcommand{\arraystretch}{1.25}
 \begin{table*}[th!]
 \centering
 \small
 \begin{tabular}{llccccccc}
 \toprule
 Dataset & Model                     & EM & Precision & Recall & F1 & Rouge-L & METEOR &  \textsc{BERTScore}  \\
         \midrule
\multirow{2}{*}{Natural Questions} 
         & \textsc{GPT-3.5 Turbo} (No ICL)  & 4.42 & 15.33 & 51.58 & 20.45 & 19.88 & 27.53 & 84.01 \\
         & \llamachat{} (No ICL)            & 8.62 & 15.66 & 26.34 & 16.06 & 16.30 & 14.44 & 83.39 \\ 
         \midrule
 \multirow{2}{*}{HotpotQA} 
         & \textsc{GPT-3.5 Turbo} (No ICL)  & 6.01 & 16.52 & 49.51 & 20.83 & 20.41 & 26.05 & 84.12 \\
         & \llamachat{} (No ICL)            & 5.27 & 10.19 & 15.39 & 10.42 & 10.58 & \phantom{0}9.50 & 81.29 \\
 \midrule
 \multirow{2}{*}{TopiOCQA} 
         & GPT-3.5 Turbo (No ICL)       & 0.48 & 11.51 & 15.09 & 10.81 & 11.07 & 13.30 & 83.69 \\
         & \llamachat{} (No ICL)        & 4.53 & 11.05 & 10.45 & \phantom{0}8.98 & \phantom{0}9.12 & \phantom{0}6.93 & 80.65 \\
 \bottomrule
 \end{tabular}

 \caption{Question-Answer performance when \emph{no context} is given to the model (only a question).}

 \label{tab:correctness_noICL}
 \end{table*}

\section{Further details on baselines}
\textbf{\label{sec:app-baselines}}
\paragraph{GPT Details.}
We set temperature and presence and frequency penalty to~$0$, $\textsc{top-p}$ to~$1.0$ and $n$ to~$1$.
Please refer to OpenAI documentation\footnote{\url{https://www.openai.com/docs/}} to learn more about these parameters.
To generate GPT answers, we prompt GPT with the following.

\begin{tcolorbox}[enhanced,attach boxed title to top center={yshift=-3mm,yshifttext=-1mm},
  colback=blue!5!white,colframe=blue!20!gray,colbacktitle=blue!20!gray,
  title=GPT Prompt,fonttitle=\bfseries,
  boxed title style={size=small,colframe=blue!20!gray} ]

        You are a helpful assistant who is able to generate brief and correct answers to questions, grounded on a given text.
        
        You are now given a "question" and a "context" possibly containing the answer to the question. Answer the question based only on the context given.
        
        If the answer to the question is not in the context, then say UNANSWERABLE. Your answer must be concise and to the point.\\

        Question: \{question\}\\
        Context: \{context\}\\
        Answer:\\

\end{tcolorbox}

\paragraph{LoRA Fine-tuning Details. }

We fine-tune the \llama{} model using LoRA adaptation~\cite{hu2021lora}.
To get a comparable number of trainable parameters as for \ourmodel, we allow \textsc{LoRA} to modify all three attention projection matrices (for the queries, keys, and values), and set $r$ to~360, $\alpha$ to~360, and dropout ratio to~0.5.
Note that all models, unless otherwise specified, are trained on the same training data that is a pooling of all five datasets discussed in Section~\ref{sec:training_dataset}.
The prompt for this fine-tuned model follows.

\begin{tcolorbox}[enhanced,attach boxed title to top center={yshift=-3mm,yshifttext=-1mm},
  colback=blue!5!white,colframe=blue!20!gray,colbacktitle=blue!20!gray,
  title=\llama-7B (LoRA) Prompt,fonttitle=\bfseries,
  boxed title style={size=small,colframe=blue!20!gray} ]

        <|system|>\\
        \{context\}\\
        <|user|>\\
        \{question\}\\
        <|assistant|>\\

\end{tcolorbox}

\paragraph{\llama-Chat Details.}

We initially fine-tuned both \llama{} and \llamachat{} (a version of \llama{} fine-tuned for chat) to our dataset, but report results on \llamachat{} as it was the model with superior performance. 
The prompt varies slightly between the version that includes the $\context$ (\llamachat{}  Prompt) and the version that excludes it (\llamachat (No ICL) Prompt).
Both prompt versions follow.

\begin{tcolorbox}[enhanced,attach boxed title to top center={yshift=-3mm,yshifttext=-1mm},
  colback=blue!5!white,colframe=blue!20!gray,colbacktitle=blue!20!gray,
  title=\llamachat{} Prompt,fonttitle=\bfseries,
  boxed title style={size=small,colframe=blue!20!gray} ]

        {[}INST{]} {<}{<}SYS{>}{>} \\
        Please answer the following question given the following passages. Please be brief. If you cannot answer the question, please reply with 'UNANSWERABLE'.\\
        \\
        {<}{<}/SYS{>}{>}\\
        \\
        \{context\}\\
        Question: \{question\}\\
        {[}/INST{]}\\
        Answer: 

\end{tcolorbox}

\begin{tcolorbox}[enhanced,attach boxed title to top center={yshift=-3mm,yshifttext=-1mm},
  colback=blue!5!white,colframe=blue!20!gray,colbacktitle=blue!20!gray,
  title=\llamachat{} (No ICL) Prompt,fonttitle=\bfseries,
  boxed title style={size=small,colframe=blue!20!gray} ]

        {[}INST{]} {<}{<}SYS{>}{>} \\
        Please answer the following question. Please be brief. If you cannot answer the question, please reply with 'UNANSWERABLE'.\\
        \\
        {<}{<}/SYS{>}{>}\\
        \\
        Question: \{question\}\\
        {[}/INST{]}\\
        Answer: 

\end{tcolorbox}

\vskip0.5cm

\section{Sensitivity analysis: position of the reference context}
\textbf{\label{sec:app-sensitivity}}
We explore the sensitivity of the models to the positioning of the reference context (useful for answering the query) within the full the context given to the model. 
Past work has shown that standard architectures based self-attention and used for language modeling tend to focus on specific parts of the past content when generating a token, and suffer from primacy and recency bias~\citep{liu2024lost,peysakhovich2023attention,xiao2023efficient}. 
To explore the performance of our models against such position shifts, we turn our focus to the \hotpotqa{} dataset, whose contexts are given by lists of (roughly) 10 sentences, out of which two are marked as \emph{useful} to answering the posed question. 
We generate four variants of the dataset that differ by the positions of the useful contexts at the beginning of these context paragraphs, in the middle, at the end, and a variant in which the two useful contexts are far away from each other and placed at the first and last positions. 
We test both our models and \llamachat{} (LoRA) on each of the datasets and see that all models are somewhat sensitive to the position of the useful contexts, but ranges are relatively small (\ie{} less than 5\%)as compared to more drastic scenarios such as what is reported in the literature for ICL (cf. Figure 1 in~\citep{liu2024lost} with ranges greater than 20\%), suggesting that fine-tuned context-conditional language models are less biased to certain parts of the context.

\renewcommand{\arraystretch}{1.25}
\begin{table}[th!]
 \centering
 \small
 \begin{tabular}{l|c|cccc}
 \toprule
       & \multicolumn{5}{|c}{Positionis of the useful contexts} \\ \\
       &                       Shuffled & Beginning & Middle &  End & Far apart \\
 Model &                            &  0,1 & 4,5 & 8,9 & 0,9   \\
 \midrule
           \llamachat{}                &  \underline{29.63} &  \textbf{33.88} &  30.45 & 32.80 & 32.05 \\
           \llama{} (LoRA)             &  \underline{71.97} &  72.28 &  71.83 & \textbf{74.35} & 73.14 \\
           \ourmodel{} (Ours)          &  \underline{43.94} &  43.52 &  \textbf{44.30} & 42.30 & 43.70 \\
           \ourmodelencoder{} (Ours)   &  \underline{54.57} &  \textbf{57.65} &  56.79 & 54.61 & 52.98 \\
 \bottomrule
 \end{tabular}

 \caption{Results on HotpotQA when we vary the position of the two useful contexts (out of a total of roughly 10 contexts per sample). We test beginning (positions 0, 1), middle (positions 4, 5), end (positions 8, 9) positions, and the case where the two contexts are far apart (positions 0, 9).}

\label{tab:results_hotpot}
\end{table}
\section{Training hyperparameters}
\textbf{\label{sec:app-hyperparams}}

The main training hyperparameters used for \ourmodel{} are reported in Table~\ref{tab:hyperparams}.

\begin{table}[ht]
\small
\centering
\begin{tabular}{lc}
\toprule
Optimizer                                 & \textsc{AdamW}                 \\
Base learning rate                        & 0.0002                \\
Learning rate warmup steps                & 2,500                  \\
Maximum gradient norm                     & 1                     \\
Batch size                                & 256                   \\
Adam $\beta_1$                            & 0.9                   \\
Adam $\beta_2$                            & 0.95                  \\
Adam $\epsilon$                           & 0.000001              \\
Weight decay                              & 0.001                 \\
Cross-attention bias                      & False                 \\
Cross-attention dropout probability       & 0.2                   \\
Cross-attention final layer               & True                  \\
Cross-attention hidden size               & 2,048                 \\
Number of Cross-attention layers          & 5                     \\
Cross-attention layer skips               & 6                     \\
Cross-attention number of attention heads & 32                    \\
Cross-attention number of key value heads & 32                    \\
Numerical precision                       & 16-bit (BF16)         \\
Number of training steps                  & 40,000                \\ \bottomrule
\end{tabular}
\caption{Training hyperparameters for \ourmodel{}.}
\label{tab:hyperparams}
\end{table}

\end{document}